\documentclass[10pt,twocolumn,letterpaper]{article}

\usepackage{iccv}
\usepackage{times}
\usepackage{epsfig}
\usepackage{graphicx}
\usepackage{amsmath}
\usepackage{amssymb}

\usepackage{graphicx}
\usepackage{amsmath}
\usepackage{amssymb}
\usepackage{booktabs}

\usepackage{graphicx}
\usepackage{amsmath}
\usepackage{amssymb}
\usepackage{booktabs}
\usepackage{threeparttable}
\usepackage{float}
\usepackage{verbatim}
\usepackage[dvipsnames]{xcolor}
\usepackage{comment}
\usepackage{algorithm}
\usepackage{colortbl}
\usepackage[pagebackref,breaklinks,colorlinks]{hyperref}
\usepackage{diagbox}
\usepackage[accsupp]{axessibility} 
\usepackage{amsmath}
\usepackage{algpseudocode}
\definecolor{darkseagreen}{rgb}{0.56, 0.74, 0.56}
\definecolor{lightpink}{rgb}{1.0, 0.71, 0.76}

\usepackage[capitalize]{cleveref}
\crefname{section}{Sec.}{Secs.}
\Crefname{section}{Section}{Sections}
\Crefname{table}{Table}{Tables}
\crefname{table}{Tab.}{Tabs.}

\usepackage{multirow}

\iccvfinalcopy 

\def\iccvPaperID{****} 

\ificcvfinal\fi

\begin{document}

\title{ScribbleSeg:~Scribble-based Interactive Image Segmentation }
\author{Xi Chen$^{1}$\quad Yau Shing Jonathan Cheung$^{1}$\quad Ser-Nam Lim$^{2}$\quad Hengshuang Zhao$^{1}$\\
$^{1}$The University of Hong Kong \quad $^{2}$Meta AI\\}

\maketitle

\begin{abstract}
Interactive segmentation enables users to extract masks by providing simple annotations to indicate the target, such as boxes, clicks, or scribbles. Among these interaction formats, scribbles are the most flexible as they can be of arbitrary shapes and sizes. This enables scribbles to provide more indications of the target object.  However, previous works mainly focus on click-based configuration, and the scribble-based setting is rarely explored. 
In this work, we attempt to formulate a standard protocol for scribble-based interactive segmentation. Basically, we design diversified strategies to simulate scribbles for training, propose a deterministic scribble generator for evaluation, 
and construct a challenging benchmark. Besides, we build a strong framework \textbf{ScribbleSeg}, consisting of a Prototype Adaption Module~(PAM) and a Corrective Refine Module~(CRM), for the task. Extensive experiments show that ScribbleSeg performs notably better than previous click-based methods. We hope this could serve as a more powerful and general solution for interactive segmentation. Our code will be made available.
\end{abstract}
\section{Introduction}

Interactive image segmentation requires users to indicate the target by providing simple annotations such as boxes, scribbles, and clicks. Compared with traditional annotation tools like the lasso or brush, interactive models could largely reduce the time and cost of creating masks, which is especially important in the era of big data.  

\begin{figure}[t]
\newcommand{\image}{\includegraphics[width=0.31\columnwidth]}
\centering 
\tabcolsep=0.05cm
\renewcommand{\arraystretch}{0.06}
\begin{tabular}{ccc}
\vspace{1mm}
\image{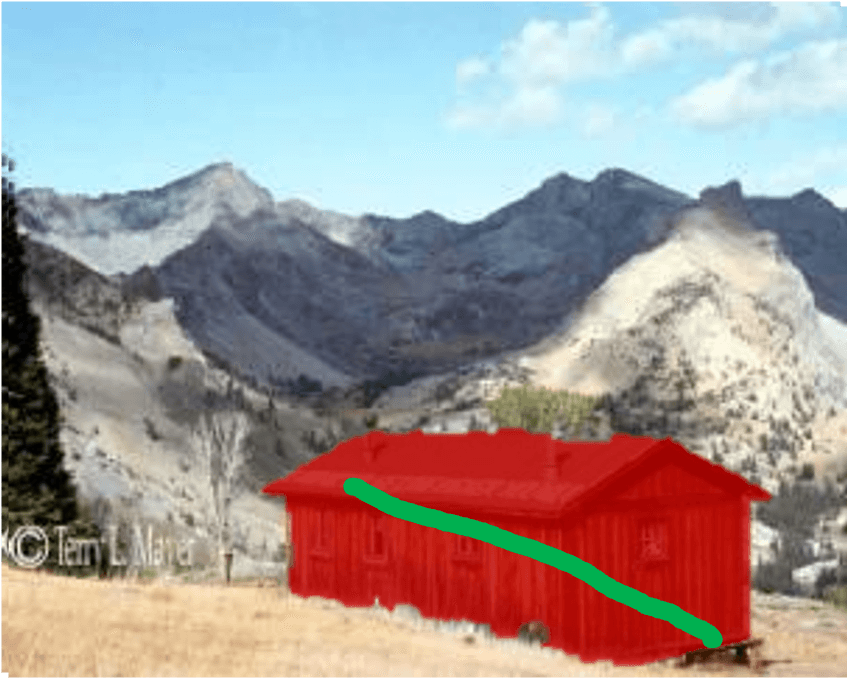} &
\image{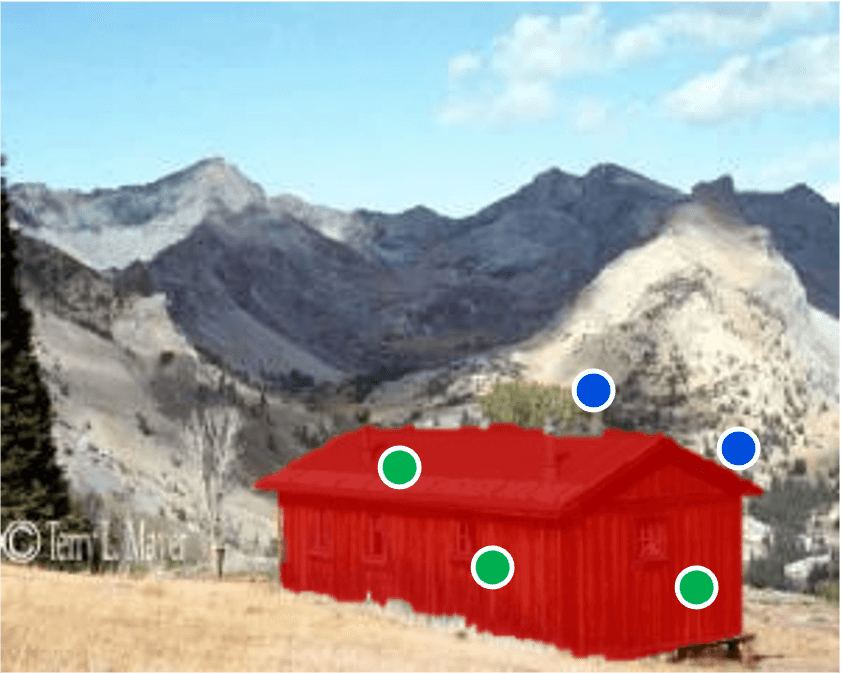} &
\image{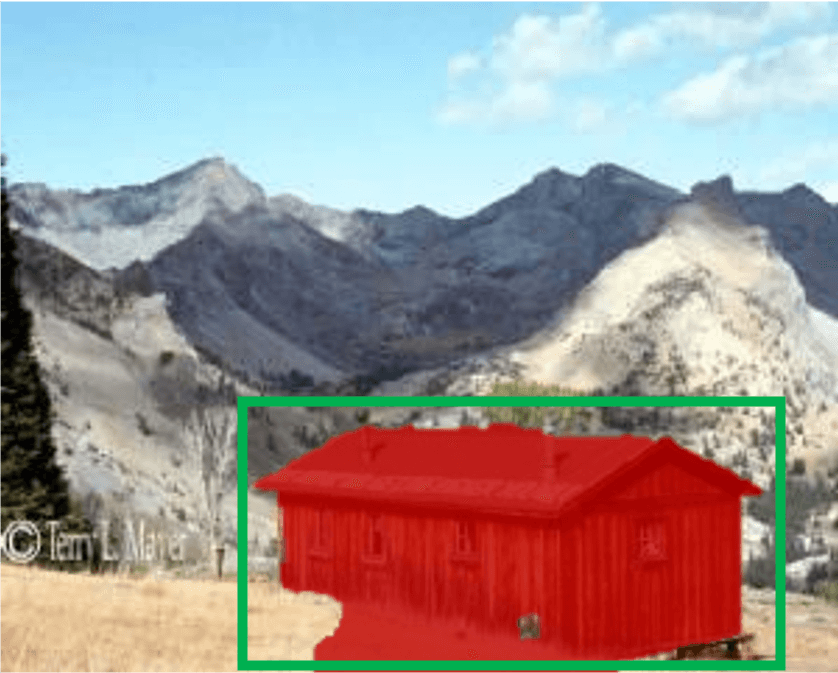} \\
\vspace{3mm}
{\footnotesize (a)~Scribble } & {\footnotesize (b)~Click } & {\footnotesize (c)~Box} \\
\end{tabular}
\caption{Comparisons of different interaction forms for interactive image segmentation. Foreground scribbles/clicks are marked in green and background scribbles/clicks in blue. }
\vspace{-3mm}
\label{fig:forms}
\end{figure}

Demonstrations of common forms of interactions are shown in Fig.~\ref{fig:forms}. Among them, we claim that drawing scribbles is the most flexible and practical way to indicate the foreground and background regions. As shown in Fig.~\ref{fig:forms} (c), although boxes could indicate the size and rough location of the target, they could not make further indications inside the rectangle. As in (b), though clicks could provide in-depth annotations, a small number of clicks is unable to indicate the shape and size of the object accurately. Thus, click-based models often require extensive interactions, especially when annotating large and complicated objects. Scribbles, on the other hand, have the combined advantages of boxes and clicks. Long scribbles can accurately indicate the shape and size of the target, while short scribbles can make detailed corrections. Scribbles are therefore regarded as an extension of clicks as they encode more information about the user's intention.

Although drawing scribbles is more practical and favorable, this topic is rarely discussed by researchers. The settings for the few existing works~\cite{bai2014error,appearancesimilarity,DeepIGeoS,IFIS} vary greatly. They use different dataset, scribble-simulation methods, evaluation metrics, and does not provide code. This makes it hard to make comparisons and hinders the development of scribble-based interactive segmentation. 
In contrast, click-based interactive segmentation is flourishing with booming works~\cite{DIOS, li2018latentdiversity, mahadevan2018iteratively, sofiiuk2021ritm, firstclick, jang2019brs,fbrs,chen2021cdnet, focalclick}.
We believe a core reason is that DIOS~\cite{DIOS}~(CVPR'16) formulated a training pipeline and evaluation protocol, thus other researchers could follow the standard setting and focus on specific points to improve the performance.   

In this work, we attempt to reference the successes of click-based methods to reformulate the task of scribble-based interactive segmentation. However, there exist gaps between these two tasks, and in our exploration, we tackle the following challenges: 

\vspace{-5mm}
\paragraph{How to get diversified scribbles for training? }
Clicks could simply be represented by a pair of coordinates, but scribbles have arbitrary shapes and complicated representations.  Previous scribble-based works majorly use feature similarities to guide the segmentation, thus they~\cite{bai2014error,appearancesimilarity,DeepIGeoS} do not need training scribbles. \cite{IFIS} simply links randomly sampled points to simulate scribbles, but is too naive to cover different real-world user interactions. 

In this paper, we design multiple meta-simulators to simulate different kinds of annotating behavior and make compositions to ensure the diversity of training samplers. We also
adapt the iterative training strategy~\cite{mahadevan2018iteratively} to add scribbles on the error regions of the last prediction.

\vspace{-4mm}
\paragraph{How to fairly evaluate the model?} 

During evaluation, we design a deterministic simulator to generate the scribble 
according to the shape and size of the given mask. We use this method to add positive/negative scribbles automatically on the max difference region between the ground truth and the predicted masks. Thus, similar to click-based settings, we could measure the Number of Interactions required to reach the target IOU. This protocol provides a unified benchmark for different types of interactions, and enables us to compare the performance among models with different interaction forms. Besides, we construct a benchmark based on ADE20K~\cite{ade20k} to evaluate the model's ability in diversified scenarios and categories.

\begin{figure}[t]
\newcommand{\image}{\includegraphics[width=0.94\columnwidth]}
\centering 
\image{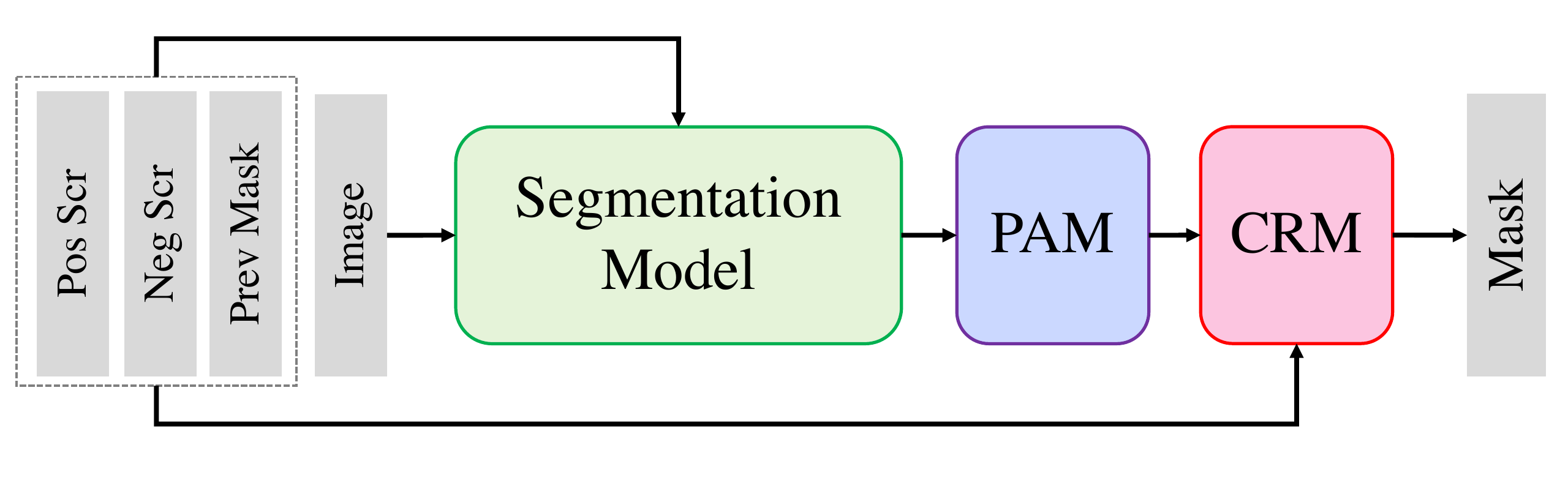} 
\vspace{-3mm}
\caption{The pipeline of ScribbleSeg. `PAM' denotes Prototype Adaption Module, and `CRM' means Corrective Refine Module.}
\label{fig:small_pipe}
\vspace{-5mm}
\end{figure}

\vspace{-5mm}
\paragraph{How to fully utilize the indications in scribbles? } Scribbles could be regarded as an extension of clicks, thus many designs from click-based methods could be transferred.  Differently, scribbles contain more indications than clicks that could be explored.
Thus, we first build a vanilla pipeline adapting from click-based methods and make specific designs considering the characteristics of scribbles. 
As in Fig.~\ref{fig:small_pipe}.  We first represent positive and negative scribbles into two binary masks. Then, we feed the 3-channel image, along with two scribble masks, and the previous prediction masks into a segmentation model. This method could be regarded as a vanilla solution. 
Starting from this baseline, we add two components to further improve its performance.

As scribbles cover more pixels than clicks, they could not only provide the location priors but also the appearance indications~(the scribble-covered regions),  
thus, we develop a Prototype Adaption Module~(PAM) to update the final projection kernel according to the user-provided scribbles. 
Besides, to produce high-quality masks, we design a Corrective Refine Module~(CRM), which takes the prediction of the segmentation model as input to estimate the probable error region and make corrections for the details. 

Our contribution could be summarized in three folds:
1)~We reformulate the task of scribble-based interactive segmentation and provide a standard train/validation protocol and benchmark. 2)~We propose ScribbleSeg, which shows strong performance for scribble-based interactive segmentation. 3)~We design PAM and CRM, which are simple and effective modules for interactive segmentation.
\vspace{-2mm}
\section{Related Work}
\vspace{-2mm}
\paragraph{Interactive image segmentation with click.}
Click-based interactive segmentation methods aim to obtain masks of the targeted objects with reference to user-provided clicks. Early methods \cite{grady2006random, boykov2001interactive,gulshan2010geodesic, kim2010nonparametric} focused on optimization-based solutions. DIOS \cite{DIOS} was the first deep learning method that proposed embedding positive and negative clicks into distance maps, then stacking them together with the image as input to the network. BRS \cite{jang2019brs} proposed an online optimization scheme for interactive segmentation, and f-BRS \cite{fbrs} sped it up by optimizing only the auxiliary variables of the network.  Later methods \cite{chen2021cdnet, firstclick, PseudoClick} have also employed a similar model architecture and provided further improvements. RITM \cite{sofiiuk2021ritm} improved model performance by taking the previous mask along with the click maps and image as input. FocalClick \cite{focalclick} performed prediction and update in localized areas, and has improved the model's efficiency and mask refinement performance.  These works all follow the train/val protocol proposed by DIOS~\cite{DIOS}, and report the performance with the same metrics,  
thus a good research community is formed.
However, the biggest disadvantage of click-based methods is that clicks embed little information, and the model, therefore, requires extensive annotations to segment objects with complicated shapes.

\vspace{-5mm}
\paragraph{Interactive image segmentation with scribbles.}
Compared to click-based segmentation, scribble-based interactive segmentation has a lot fewer methods proposed. 

Early works~\cite{ScribbleSup, GRanking,bai2014error,GFilter} used graph constraints, energy functions, or Gabor filters to deal with scribbles. 
DeepIGeoS \cite{DeepIGeoS} encoded scribbles with geodesic distance transforms and performed mask refinement with it. \cite{IFIS} allowed the sharing of scribble annotations across multiple object regions.  \cite{appearancesimilarity} leveraged the appearance similarity to propagate scribble information to other regions. In the current field, however, there has yet to be a standard training and validation protocol proposed. Researchers would use IOU~\cite{IFIS}, Dice Coefficient~\cite{bai2014error}, and annotating time~\cite{appearancesimilarity} as metrics and report evaluation result on different benchmarks.

\begin{figure*}[t]
\newcommand{\image}{\includegraphics[width=1.98\columnwidth]}
\centering 
\image{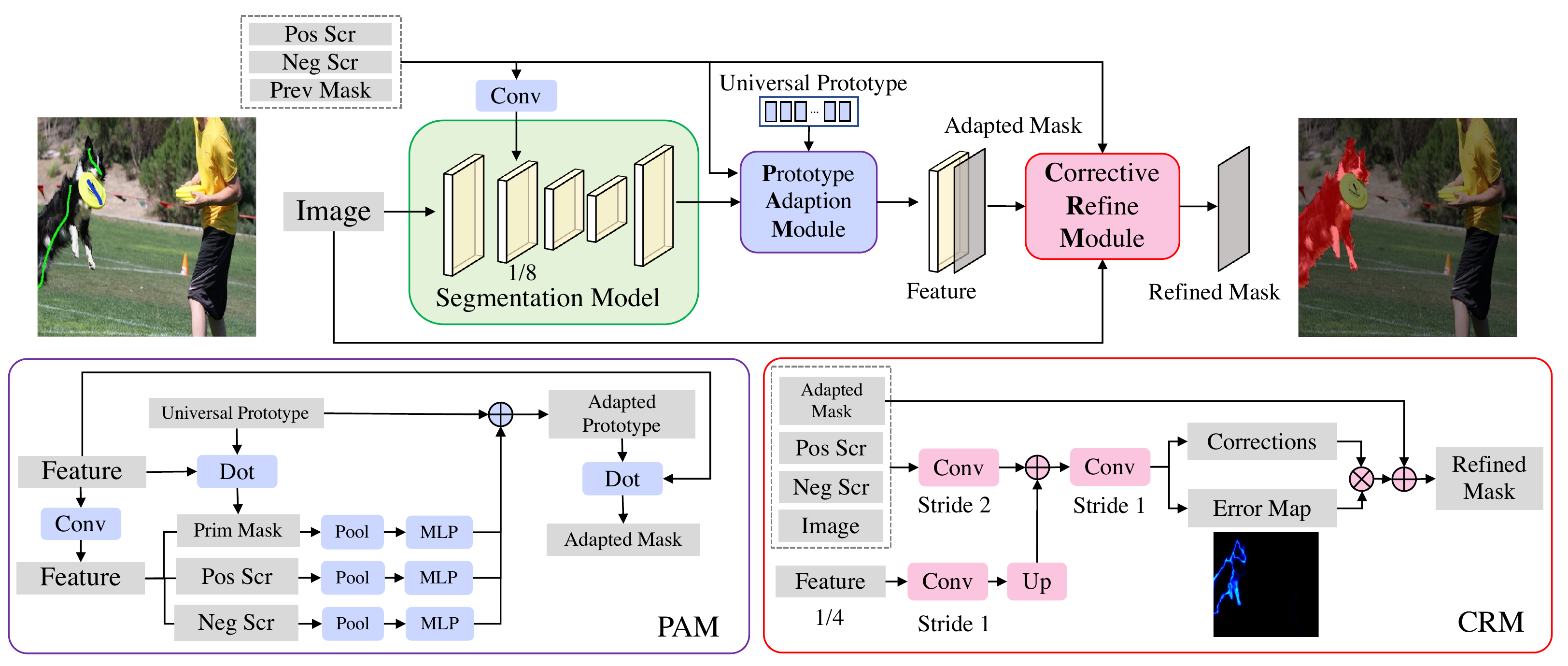} 
\vspace{-2pt}
\caption{The demonstration for the pipeline of ScribbleSeg. We take the image, the scribble maps, and the previous mask as inputs to extract the mask of the target object. The positive and negative scribbles are marked in green and blue.
\textbf{PAM} denotes Prototype Adaption Module. \textbf{CRM} means Corrective Refine Module. The detailed structure of KAM and CRM can be found at the bottom part of the figure.}
\label{fig:pipeline}
\vspace{-4mm}
\end{figure*}

\vspace{-5mm}
\paragraph{Scribble-based video object segmentation.}
DAVIS-2018~\cite{davisinteractive} provide a track for scribble-based video object segmentation, which aims to produce masks for object annotated in all frames of a video, and use scribbles to make target indications and corrections. Although using scribbles to refine masks is an important step in this task, DAVIS-2018~\cite{davisinteractive} only cares about the performance on the whole video sequence, and therefore previous works~\cite{Nagaraja2015VideoSW,FastVOS,GIVOS,IVOS,MIVO} often use a simple module to deal with scribbles and focus on the information propagation between frames.
\section{Method}

We first give an overall introduction for the task of scribble-based interactive segmentation and survey the current status of the community in Sec.~\ref{overall}. Then, we elaborate on the detailed model structure of ScribbleSeg in Sec.~\ref{framework}. Afterward, we introduce the training protocol in Sec.~\ref{sec:train}. In Sec.~\ref{val_protocol}, we describe the evaluation method and our constructed benchmark for evaluation.  

\subsection{Task Overview}
\label{overall}
\paragraph{Task definition.}
To extract a target mask, the user sequentially draws positive/negative scribbles to add/remove mask regions, and the model returns the predicted mask after receiving new interactions. For each interaction period, only one additional scribble could be provided. 

\vspace{-3mm}
\paragraph{Exisiting works.}
We survey the  scribble-based interactive segmentation methods in Tab.~\ref{tab:survey}. Most of the existing works use different protocols~(domain, data, metrics), which makes it hard to make comparisons and hinders the development of the community. However, as scribbles contain more information than clicks, it has the potential to be a more promising choice for interactive segmentation.  Thus, the community needs a standard framework to unleash the potential of scribble-based interactive segmentation.

\begin{table}[h]
\small
\begin{center}
\scalebox{0.7}{
\begin{threeparttable}
\begin{tabular}{l|c|c|c|c|c }
\toprule[1pt]
    &  Train Data  & Train Scribbles &  Test Data & Test Scribbles  & Metrics  \\
\hline
\cite{IFIS} & COCO & Linked points  & COCO & Linked points & IoU per Scribble    \\
\cite{appearancesimilarity} & COCO & None & COCO & Human Drawn   & Annotation Time  \\
\cite{bai2014error} & None  & None & GrabCut & Random Pixels & Label Accuracy   \\
\cite{DeepIGeoS} & Medical & None & Medical & Random Pixels & Dice Score  \\

\bottomrule[1pt]
\end{tabular}
\end{threeparttable}
}
\end{center}
\vspace{-2mm}
\caption{ Survey for existing works of scribble-based interactive segmentation. Their configurations differ from each other.
}
\label{tab:survey}
\end{table}

\subsection{Framework for ScribbleSeg}
\label{framework}
As shown in Fig.~\ref{fig:pipeline}, we first feed the image and interaction maps into a
segmentation model. This is a commonly used baseline solution transferred from click-based methods~\cite{sofiiuk2021ritm}. To improve its performance, we analyze the characteristics of using scribbles as interactions to design novel components. Accordingly, we develop two modules: 

\vspace{-4mm}
\paragraph{Prototype Adaption.} 
Scribbles often cover more pixels than clicks, thus besides providing position/shape priors like clicks, the scribble-covered regions could better indicate the target appearance and semantics. According to this property, we propose to use the scribble-covered regions to enhance the target extraction procedure.     

Masks extraction could be understood as a correlation between the projected features and  learned prototypes~(the last FC layer for the segmentation model).  Traditional semantic segmentation models learn fixed prototypes for each category. However, interactive segmentation models learn only one prototype for the universal targets.  As the segmentation target could be any object, stuff, or part, it is challenging to use a single prototype to represent the diverse target.

We propose the Prototype Adaption Module~(PAM), which dynamically adapts the universal prototype~(the last projection kernel) by interacting with the scribble-covered features. Thus, the parameters of the segmentation model become image-specific and scribble-specific.

As shown in the left bottom part of Fig.~\ref{fig:pipeline}, we first use the universal prototype as the convolution kernel to generate a primitive mask $\mathbf{M}_{prim}$ via dot production with the final feature map. This prototype is  initialized with learnable parameters. Afterward, we use the two user-provided scribble maps, and the primitive mask prediction to pool the feature map into three embeddings. Then, we project those embeddings using MLP layers and add them to the original prototype. As the labels of the scribble-marked regions are known, they contain more cues that indicate the user's intentions. The primitive mask could also give a global representation for the segmentation target, which helps construct a dynamic prototype with high-consistency representations.  Finally, we use this adapted prototype to predict the adapted mask, and note it as $\mathbf{M}_{ada}$.

\vspace{-2mm}
\paragraph{Corrective Refine.}
The scribbles could also provide indications for refining the segmentation details.
We propose Corrective Refine Module~(CRM) to make modifications on the $\mathbf{M}_{ada}$ predicted by PAM. As demonstrated in the right bottom of Fig.~\ref{fig:pipeline}, the first branch of CRM concatenates the predicted mask with the scribble maps and the original image to extract the features with fine details. The second branch fuses the detached features from the segmentation model. The features in CRM are kept at the resolution of stride-2 to preserve the fine details. Afterwards, we predict an error map $\mathbf{M}_{error}$ and a correction map $\mathbf{M}_{corr}$. The error map is supervised with the difference between the ground truth and the $\mathbf{M}_{ada}$. 
The final prediction is a combination of the $\mathbf{M}_{ada}$ and $\mathbf{M}_{corr}$ in the predicted error region $\mathbf{M}_{error}$. The structure of CRM is inspired by the refiner of FocalClick~\cite{focalclick}. The core differences are, we detach the feature of the segmentation model, and we use the error map to guide the detail correction process. The effectiveness of these modifications would be verified in the experiments. 

\vspace{-2mm}
\paragraph{Training losses.} The principle supervision is for the final refined prediction, for which we use normalized focal loss~(NFL)~\cite{sofiiuk2021ritm}, and note it as $\mathcal{L}_{ref}$. Besides, we also use NFL to supervise the primitive and adapted masks produced by KAM, and note them as $\mathcal{L}_{prim}$, and $\mathcal{L}_{ada}$. In addition, the error map of CRM is supervised with binary cross-entropy loss, $\mathcal{L}_{error}$. The total loss is a combination of these losses, where $\alpha, \beta, \gamma$ all equal 0.4.
\begin{equation}
    \mathcal{L}_{total} = \mathcal{L}_{ref} +  \alpha \mathcal{L}_{prim} + \beta \mathcal{L}_{ada} + \gamma \mathcal{L}_{error}
\label{equ:loss}
\end{equation}

\begin{figure}[t]
\newcommand{\image}{\includegraphics[width=0.32\columnwidth]}
\centering 
\tabcolsep=0.04cm
\renewcommand{\arraystretch}{0.06}
\begin{tabular}{ccc}
\vspace{3pt}
\image{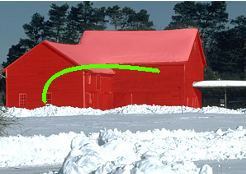} &
\image{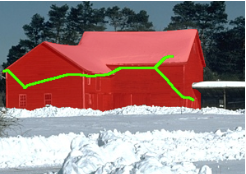} &
\image{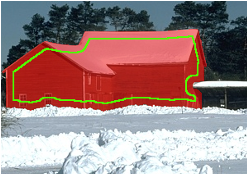} \\
\vspace{3pt}
{\footnotesize (a)~Bezier Scribble } & {\footnotesize (b)~Axial Scribble} & {\footnotesize (c)~Boundary Scribble} \\

\vspace{3pt}
\image{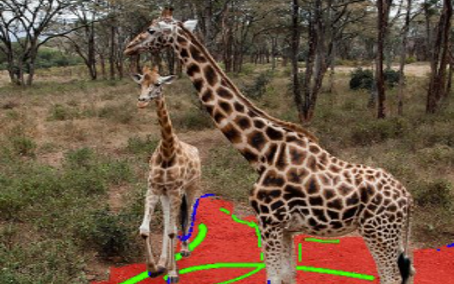} &
\image{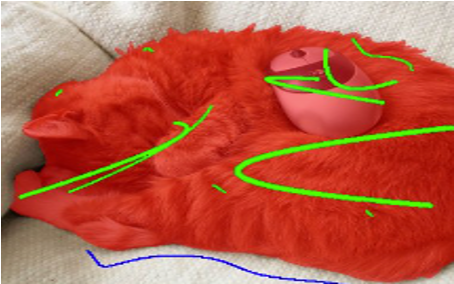} &
\image{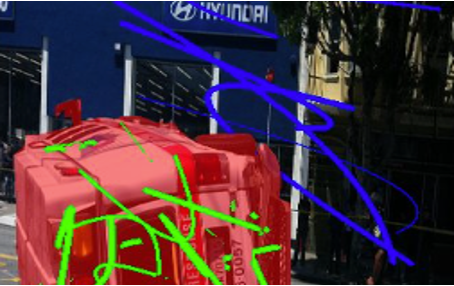} \\
\multicolumn{3}{c}{ \footnotesize{(d)~Random compositions of scribbles during training.}} \\
\end{tabular}
\vspace{0mm}
\caption{Demonstrations of the meta scribble simulators. We combine these strategies to generate  scribbles during training. }
\vspace{-2mm}
\label{fig:scribbles_train}
\end{figure}

\subsection{Training Scribble Simulation}
\label{sec:train}

\paragraph{Meta-simulators.} The naive methods~\cite{IFIS,bai2014error,DeepIGeoS} use random dilated points to simulate scribbles, which could not satisfy the diversity. Additionally, we 
develop multiple meta-simulators to generate various scribbles.  As demonstrated in Fig.~\ref{fig:scribbles_train}, the bezier scribble uses bezier function to draw curves within the mask regions; the axial scribble calculates the media axis of the given mask; the boundary scribble draws lines along with the mask boundary. For the stroke thickness, we randomly choose values from 3 to 7.

\vspace{-4mm}
\paragraph{Scribble composition.} We combine these four strategies to generate diversified scribbles during training with carefully tuned ratios, some of the results could be found in Fig.~\ref{fig:scribbles_train} (d). When starting from a void previous mask, we generate positive scribbles in the foreground regions, and negative scribbles in the background.  

\vspace{-4mm}
\paragraph{Iterative sampling.} To better simulate the practical usage, we add scribbles iteratively. When starting from a previous mask, we first calculate the False Positive~(FP) and False Negative~(FN) regions and generate negative and positive scribbles accordingly. 
We develop two strategies to simulate the flawed masks. The first one is applying the iterative training schema~\cite{mahadevan2018iteratively}. 
Another strategy is to exert random perturbations on the ground truth masks, where we use random dilation, erosion, translation, and local erasing.  We combine these two strategies during training and make experiments to find the best combination ratio.

\begin{figure}[t]
\newcommand{\image}{\includegraphics[width=1\columnwidth]}
\centering 
\image{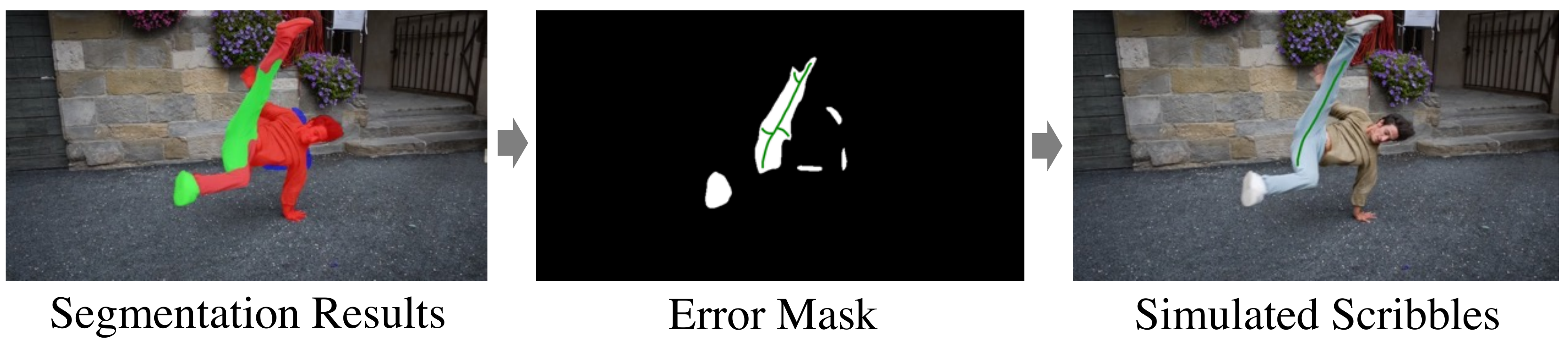}
\vspace{-4mm}
\caption{The procedure of deterministic scribble generation. The true positives, false negatives, and false positives of the segmentation result are marked in red, green, and blue respectively. }
\label{fig:scribble_protocol}
\vspace{-4mm}
\end{figure}

\begin{algorithm}[t] 
\caption{Deterministic Scribble Simulator}
\label{code}
\begin{algorithmic}[1]
    \small
    \State $max\_mask = \max(error\_mask)$
    \State $skel\_mask = \hspace{0.1cm} $ \Call{MedialAxis}{$max\_mask$}
    \State $Graph = \hspace{0.1cm} $\Call{RadiusNeighbourGraph}{$skel\_mask$}
    \For {$subgraph \in \hspace{0.1cm} $\Call{Connected}{$Graph$}}
        \While {true}
            \State $cycle = \hspace{0.1cm} $\Call{FindCycle}{$subgraph$}
            \If {$cycle == \text{None}$} $\text{break}$
            \Else \hspace{0.1cm}$ 
            $\Call{RemoveCycle}{$subgraph$, $cycle$}
            \EndIf

        \EndWhile
    \EndFor
    
    \State $distance = [~]$
    \For {$v \in Graph$.$nodes$()}
        \State $max\_path = \hspace{0.1cm} $\Call{ShortestPath}{$Graph$, $v$}
        \State $distance$.append($max\_path)$
    
    \EndFor
    \State $longest\_path = \max(distance)$
    \State $scribble = \hspace{0.1cm} $\Call{BezierCurve}{$longest\_path$}
\end{algorithmic}
\end{algorithm}

\subsection{Evaluation Method}
\vspace{-5pt}
\label{val_protocol}
In this section, we introduce our evaluation protocol that could compare different methods fairly and automatically.

\vspace{-4mm}
\paragraph{Revisiting click-based evaluation.}
DIOS~\cite{DIOS} proposes the evaluation methods that followed by almost all previous click segmentation works~\cite{fbrs,jang2019brs,firstclick,sofiiuk2021ritm,sofiiuk2021ritm,chen2021cdnet,focalclick}. In this setting, the simulated clicks are added sequentially on the center of the maximum error regions for the previous prediction. For example, the first positive click is added at the center of the ground truth mask. Then, we calculate the error regions of the prediction and extract the maximum connected area. Afterwards, an additional positive/negative click would be placed at the center of this maximum connected area. Thus, clicks would be added automatically until the prediction reaches the target IOU, or when the number of clicks reaches the limitation. In this way, we could report NoC85/90~(the average Number of Clicks required to reach the IOU 85/90\% ), and NoF$^{20}$85/90~(the Numbers of Failures to reach IOU 85/90\% within 20 clicks).     

\vspace{-4mm}
\paragraph{Deterministic scribble-simulator.} We attempt to extend the click-based protocol into a more general form for scribbles. The challenge is that clicks could simply be added at the center of the error region, but not for scribbles, as they have various shapes, which introduces randomness. 

Accordingly, We develop a \textbf{deterministic} scribble simulator that could simulate human-like scribbles according to the shape and size of the given mask.
The pseudo-code for the scribble generation process is shown in Alg.~\ref{code}, and the demonstration is given in Fig.~\ref{fig:scribble_protocol}.
Similar to the click-based protocol, we first calculate the max error regions.  Then, we compute the medial axis for the largest error mask to obtain the skeleton of the objects. Afterwards, we transform the skeleton mask into a radius neighbor graph, where the neighborhood of a vertex is points at a distance less than the radius from it. Then we divide the graph into connected components sub-graphs and remove its cycles. Finally, we will create a Bezier curve with the points in the graph's longest path. The thickness of the stroke is set to a fixed value~(3 as default).

With this deterministic scribble simulator, we could iteratively add scribbles on the FP or FN regions. Thus, we generalize the NoC metric for clicks to NoI~(Number of Interactions), and report NoI85/90, NoF$^{20}$85/90.

\vspace{-4mm}
\paragraph{Evaluation benchmark.}
To perform comparisons with previous methods, we first evaluate ScribbleSeg on the benchmarks~\cite{rother2004grabcut,berkeley,SBD,davis} used by click-based models. However, they have the following disadvantages:
GrabCut~\cite{rother2004grabcut} and Berkeley~\cite{berkeley} only have 50 and 100 test images respectively, which could not provide convincing results.
SBD~\cite{SBD} contains 2802 samples, but the mask annotations are coarse, thus often causing inconsistent results, which has been discussed in ~\cite{chen2021cdnet,focalclick}.   
DAVIS~\cite{davis} contains 345 annotations, but the targets are all saliency objects. 

In general, the benchmark above only contains relatively easy cases for the salient object.
However, a good annotation tool should be able to deal with large-scope categories in complex scenes. Therefore, we additionally use ADE20K~\cite{ade20k} to evaluate our model, which covers both things and stuff for 150 categories. We use the panoptic format annotation for the ADE20K validation split. For each category, we randomly pick 5 samples~(we take the max number if there are fewer than 5 samples) and we have obtained 246 samples for stuff and 499 samples for things.

\section{Experiments}
\subsection{Experiment Configurations}
\paragraph{Implementation details.} The segmentation model in ScribbleSeg could be an arbitrary semantic segmentation network. In this work, we choose the SegFormer~\cite{xie2021segformer} for experiments following previous SOTAs~\cite{focalclick} of click-based segmentation. The input size of ScribbleSeg is kept as $384\times384$ during training and inference. 

During inference, we apply the zoom-in strategy proposed in \cite{fbrs}. Concretely, starting from the second stroke of scribble, we calculate the external box according to the previous mask and the current scribbles and expand it with a ratio of 1.4. Then, we crop the model input according to this expanded box and resize it to $384\times384$. This allows the model to focus on the target region, which is especially effective when the target is small in the image.

\vspace{-3mm}
\paragraph{Training configurations.} Following previous works of click-based segmentation~\cite{focalclick,sofiiuk2021ritm}, we train our model on the combined dataset of COCO~\cite{lin2014coco} and LVIS~\cite{gupta2019lvis}. For data augmentation, we use random flip and random resize with a scale ratio from 0.75 to 1.4. We take 3,000 images for each epoch and train ScribbleSeg with 150 epochs. The initial learning rate is set as 0.0005, and we add two lr decay with the ratio of 0.1 at the epoch of 110 and 130. For the optimizer, we pick ADAM with $\beta_{1}=0.9$ and $\beta_{2}=0.999$.

\begin{table*}[t]
\begin{center}
\scalebox{0.85}{
\begin{tabular}{l|l|c|ccc|ccc|ccc}
\toprule[1pt]
& \multicolumn{1}{l|}{} & & \multicolumn{3}{c|}{ADE-Stuff} & \multicolumn{3}{c}{ADE-Thing} & \multicolumn{3}{c}{ADE-Full}  \\

& Method & Train Interaction & NoI~85 & NoI~90 & NoF~90 & NoI~85 & NoI~90 & NoF~90 & NoI~85 & NoI~90 & NoF~90 \\
\hline
1 & AppearanceSim~\cite{appearancesimilarity} & None & 9.61  & 15.34 & 92 & 10.01  & 12.12 & 186 & 9.87 & 13.18 & 278  \\
2 & RITM-scribble~\cite{sofiiuk2021ritm} & Click points & 7.20  & 10.33 & 76 & 7.91  & 10.82 & 166 & 7.73 & 10.65 & 242  \\
3 & RITM-scribble~\cite{sofiiuk2021ritm}  & Linked points~\cite{IFIS}   & 6.11 & 7.92  & 58 & 6.99 &  8.74 & 131 & 6.69 & 8.47 & 189 \\

\hline
4 & ScribbleBase-B0 & Composed scribbles   & 4.89 & 7.14  & 41 & 5.61 &  8.19 & 111 & 5.37  & 7.84 & 152   \\
5 & ScribbleSeg-B0  & Composed scribbles   & 4.77 & 6.90  & 41 & 5.08 & 7.63  & \bf105 &  4.97  & 7.38 & \bf146   \\
\rowcolor{gray!20} 
6 & ScribbleSeg-B3  & Composed scribbles  & \bf4.41 & \bf6.57  & \bf39 & \bf5.06 & \bf7.46  & 109 & \bf4.84  & \bf7.16 & 148   \\
\bottomrule
\end{tabular}
}
\end{center}
\vspace{-2mm}
\caption{Comparison results on our proposed ADE20K benchmark. We report the performance for stuff and thing categories respectively. All models are trained on COCO+LVIS. `NoI~85/90' denotes the average Number of Interactions required the get IoU of 85/90\%. `B0/B3' denotes using SegFormer-B0/B3~\cite{xie2021segformer} as the segmentation model.  } 
\label{tab:ade}
\vspace{0mm}
\end{table*}

\subsection{ Quantitative Analysis} 
We first give a quantitative analysis for ScribbleSeg on our newly constructed ADE20K benchmark.  Using the proposed evaluation protocol, we evaluate the model performance with NoI@IoU, which means the average Number of Interactions required to reach the target IoU. 

As explained in Sec.~\ref{overall}, previous works are not suitable for comparison. Hence, we reproduce some representative baselines for comparison in Tab.~\ref{tab:ade}. In row~1, we train 
a similarity-based model like~\cite{appearancesimilarity}, which shows poor performance as it could not deal with the fine details. Row~2 corresponds to using the click-based solution~\cite{sofiiuk2021ritm} to deal with scribbles. We find that, although clicks could be regarded as short scribbles, directly using click-based models could not get satisfactory results. In row~3, we follow IFIS~\cite{IFIS} to simulate scribbles via linking randomly sampled points. This strategy brings improvements  compared to using click disks. However, it still shows a big gap compared with our work which is demonstrated in the second part.  

Row~4 denotes the baseline version without PAM and CRM, and row~5 and 6 show the full ScribbleSeg using SegFormer-B0/B3~\cite{xie2021segformer} as the segmentation model. It could be observed that ScribbleSeg achieves significantly better performance than its counterparts. 
On stuff categories, the advantage of ScribbleSeg is even larger. This is because scribbles can provide more indications compared to clicks in stuff categories that cover a relatively big region of arbitrary shapes and complicated semantics.

\begin{table}[t]
\begin{center}
\scalebox{0.7}{
\begin{tabular}{l|cc|cc|cc}
\toprule[1pt]
\multicolumn{1}{l|}{} & \multicolumn{2}{c|}{Berkeley} & \multicolumn{2}{c|}{DAVIS} & \multicolumn{2}{c}{ADE-full}  \\
Pred : Pertub   & NoI~90 & NoF~90 & NoI~90 & NoF~90 & NoI~90 & NoF~90 \\
\hline
0 : 0  & 2.30 & 3 & 6.02 & 64 & 8.95  & 179 \\
1 : 0  & 2.06 & 1 & 5.46 & 61 & 8.61  & 176 \\
1 : 0.2  & 1.79 & 0 & 5.21 & 52 & 8.01  & 156 \\
\rowcolor{gray!20} 
1 : 0.4  & 1.77 & 0 & 5.18 & 51 & 7.98  & 153 \\
1 : 0.6  & 1.81 & 0 & 5.33 & 54 & 8.12 & 160 \\

\bottomrule
\end{tabular}
}
\end{center}
\vspace{-2mm}
\caption{ Comparison results of using different ratios of predicted masks~(generated by iterative training) and perturbed ground truth mask to simulate the previous mask.  } 
\label{tab:prev_mask}
\vspace{-2mm}
\end{table}

\begin{table}[t]
\begin{center}
\scalebox{0.7}{
\begin{tabular}{l|cc|cc|cc}
\toprule[1pt]
\multicolumn{1}{l|}{} & \multicolumn{2}{c|}{Berkeley} & \multicolumn{2}{c|}{DAVIS} & \multicolumn{2}{c}{ADE-full}  \\
Max Number  & NoI~90 & NoF~90 & NoI~90 & NoF~90 & NoI~90 & NoF~90 \\
\hline
8 & 1.92 & 1 & 5.34 & 56 & 8.33  & 164 \\
12 & 1.81 & 0 & 5.17 & 51 & 8.11  & 158 \\
\rowcolor{gray!20} 
16 & 1.77 & 0 & 5.18 & 51 & 7.98  & 153 \\
20  & 1.76 & 0 & 5.23 & 52 & 8.01  & 155 \\
\bottomrule
\end{tabular}
}
\end{center}
\vspace{-2mm}
\caption{Comparisons for the maximum number of strokes for the simulated scribbles during training.} 
\label{tab:num_scr}
\vspace{-2mm}
\end{table}

\begin{table}[t]
\begin{center}
\scalebox{0.73}{
\begin{tabular}{l|cc|cc|cc}
\toprule[1pt]
\multicolumn{1}{l|}{} & \multicolumn{2}{c|}{Berkeley} & \multicolumn{2}{c|}{DAVIS} & \multicolumn{2}{c}{ADE-full}  \\
Proportions & NoI~90 & NoF~90 & NoI~90 & NoF~90 & NoI~90 & NoF~90 \\
\hline
0    & 1.74 & 0 & 5.26 & 52 & 8.12  & 155 \\
\rowcolor{gray!20} 
0.2  & 1.77 & 0 & 5.18 & 51 & 7.98  & 153 \\
0.4  & 1.79 & 0 & 5.29 & 53 & 8.09  & 153 \\
\bottomrule
\end{tabular}
}
\end{center}
\vspace{-2mm}      
\caption{We set the Bezier curve as the principle strategy for scribble simulation, and analyze the proportions for the axial and boundary scribble during training.} 
\label{tab:scribble}
\vspace{-2mm}
\end{table}

\begin{table}[t]
\begin{center}
\scalebox{0.7}{
\begin{tabular}{l|cc|cc|cc}
\toprule[1pt]
\multicolumn{1}{l|}{} & \multicolumn{2}{c|}{Berkeley} & \multicolumn{2}{c|}{DAVIS} & \multicolumn{2}{c}{ADE-full}  \\
Scribble Type  & NoI~90 & NoF~90 & NoI~90 & NoF~90 & NoI~90 & NoF~90 \\
\hline
Allow Error  & 2.01 & 1 & 5.45 & 57 & 8.98  & 183 \\
Clean Boundary  & 1.77 & 0 & 5.18 & 51 & 7.98  & 153 \\
\rowcolor{gray!20} 
Protect Boundary  & 1.66 & 0 & 5.14 & 53 & 7.84 & 152 \\
\bottomrule
\end{tabular}
}
\end{center}
\vspace{-2mm}
\caption{Different strategies for dealing with the simulated scribbles that are near the boundaries of the ground truth mask.} 
\label{tab:boudary}
\vspace{-2mm}
\end{table}

\subsection{Ablations Studies}
After verifying our promising performance, in this section, we dive into the details of our framework, including the basic settings of scribble-simulation, and the two novel components: Prototype Adaption Module~(PAM) and Corrective Refine Module~(CRM). 

We first make an analysis of the basic settings to explore what makes a strong baseline for scribble-based interactive segmentation. We use a vanilla model without PAM and CRM and mainly focus on exploring the strategies of simulating the previous masks and the scribbles during training.  

\vspace{-4mm}
\paragraph{Flawed masks simulation.} We first analyze the simulation of previous masks. In Tab.~\ref{tab:prev_mask}, we explore the combined ratio for two kinds of previous mask generation methods introduced in Sec.~\ref{sec:train}. `Pred' denotes the predicted masks of the current model, which is generated by iterative training~\cite{mahadevan2018iteratively}. `Pertub' means the perturbed mask of the ground truth. The results show that the previous masks are important for interactive segmentation, and we choose the combination ratio that achieves the best performance.

\vspace{-4mm}
\paragraph{Scribble simulation.} Afterwards, we make explorations for the scribble simulation strategies. In Tab.~\ref{tab:num_scr}, we report the performance of using different numbers of scribble strokes. During training, we set the maximum number of strokes, and randomly pick a stroke number with a probability decay of 0.8, which means that the probability of choosing $n$ strokes is 0.8 of $n-1$. 

In Tab.~\ref{tab:scribble}, we tune the combined ratio for the three types of scribbles introduced in Sec.~\ref{sec:train}. We use the Bezier curve as the principle strategy, and tune the ratios of axial and boundary scribbles. As the Bezier curve is the closest to human-drawn scribbles, a higher portion results in better performance. At the same time, a small portion of the axial and boundary scribbles could increase the training diversity, which is also beneficial.

Tab.~\ref{tab:boudary} shows that dealing with scribbles in boundary regions is also important. `Allow Error' means allowing the simulated scribbles to slightly exceed the ground truth masks. `Clean Boundary' denotes removing the exceeded parts of simulated scribbles. `Protect Boundary' describes eroding the ground truth mask as the target region to simulate scribbles, and is proven to be effective.

\begin{table}[t]
\begin{center}
\scalebox{0.7}{
\begin{tabular}{l|cc|cc|cc}
\toprule[1pt]
\multicolumn{1}{l|}{} & \multicolumn{2}{c|}{Berkeley} & \multicolumn{2}{c|}{DAVIS} & \multicolumn{2}{c}{ADE-full}  \\
Method  & NoI~90 & NoF~90 & NoI~90 & NoF~90 & NoI~90 & NoF~90 \\
\hline
StrongBase   & 1.66 & 0 & 5.14 & 53 & 7.84 & 152 \\
+PAM  & 1.68 & 0 & 4.98 & 52 & 7.52  & 152 \\
+CRM  & 1.53 & 0 & 4.76 & 51 & 7.46  & 148 \\
\rowcolor{gray!20} 
+PAM+CRM & 1.49 & 0 & 4.68 & 50 & 7.38 & 146 \\
\bottomrule
\end{tabular}
}
\end{center}
\vspace{-3mm}
\caption{Ablation studies for our novel components. } 
\label{tab:ablation}
\vspace{-3mm}
\end{table}

\begin{table}[t]
\begin{center}
\scalebox{0.7}{
\begin{tabular}{l|cc|cc|cc}
\toprule[1pt]
\multicolumn{1}{l|}{} & \multicolumn{2}{c|}{Berkeley} & \multicolumn{2}{c|}{DAVIS} & \multicolumn{2}{c}{ADE-full}  \\
Method  & NoI~90 & NoF~90 & NoI~90 & NoF~90 & NoI~90 & NoF~90 \\
\hline
StrongBase   & 1.66 & 0 & 5.14 & 53 & 7.84 & 152 \\
+Scribble Pool  & 1.66 & 0 & 5.01 & 51 & 7.60  & 151 \\
+Mask Pool  & 1.64 & 0 & 5.06 & 53 & 7.63  & 153 \\
\rowcolor{gray!20} 
+Full PAM  & 1.68 & 0 & 4.98 & 52 & 7.52  & 152 \\
\bottomrule
\end{tabular}
}
\end{center}
\vspace{-3mm}
\caption{Ablation studies for the details of PAM. } 
\label{tab:PAM}
\vspace{-3mm}
\end{table}

\begin{table}[t]
\begin{center}
\scalebox{0.65}{
\begin{tabular}{l|cc|cc|cc}
\toprule[1pt]
\multicolumn{1}{l|}{} & \multicolumn{2}{c|}{Berkeley} & \multicolumn{2}{c|}{DAVIS} & \multicolumn{2}{c}{ADE-full}  \\
Method  & NoI~90 & NoF~90 & NoI~90 & NoF~90 & NoI~90 & NoF~90 \\
\hline
StrongBase  & 1.66 & 0 & 5.14 & 53 & 7.84 & 152 \\
+Refiner~\cite{focalclick}  & 1.61 & 0 & 5.01 & 52 & 7.66 & 153 \\
+CRM w/o Detach & 1.55 & 0 & 4.95 & 52 & 7.64 & 151 \\
+CRM w/o Error Map  & 1.52 & 0 & 4.87 & 52 & 7.65 & 152 \\
+CRM w/o Scribbles  & 1.58 & 0 & 4.93 & 51 & 7.56 & 150 \\
\rowcolor{gray!20} 
+Full CRM  & 1.53 & 0 & 4.76 & 51 & 7.46 & 148 \\
\bottomrule
\end{tabular}
}
\end{center}
\vspace{-3mm}
\caption{Ablation studies for the details of CRM. } 
\label{tab:CRM}
\vspace{-2mm}
\end{table}

\begin{figure}[t]
\newcommand{\image}{\includegraphics[width=0.31\columnwidth]}
\centering 
\tabcolsep=0.02cm
\renewcommand{\arraystretch}{0.06}
\begin{tabular}{ccc}
\vspace{3pt}
\image{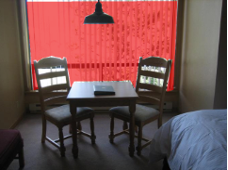} &
\image{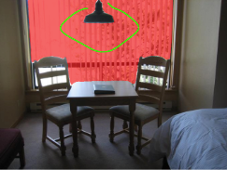} &
\image{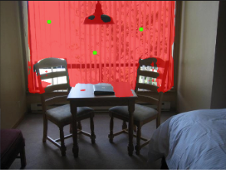} \\
\vspace{3pt}
{\footnotesize (a)~Ground Truth } &{\footnotesize (a)~1 Scr~:~92.3\% } & {\footnotesize (a)~3 Clicks~:~83.5\% } \\
\vspace{3pt}
\image{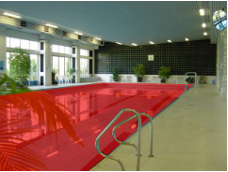} &
\image{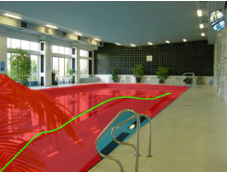} &
\image{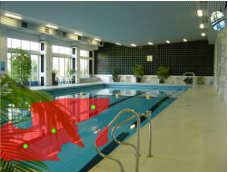} \\
\vspace{3pt}
{\footnotesize (b)~Ground Truth } &{\footnotesize (b)~1 Scr~:~86.9\% } & {\footnotesize (b)~3 Clicks~:~37.8\% } \\
\vspace{3pt}
\image{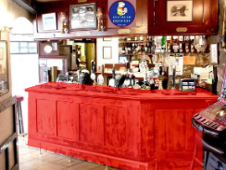} &
\image{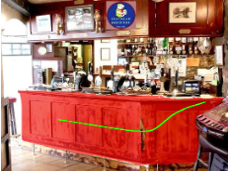} &
\image{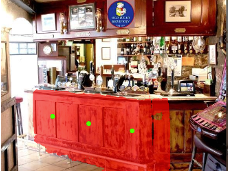} \\
\vspace{3pt}
{\footnotesize (c)~Ground Truth } &{\footnotesize (c)~1 Scr~:~76.8\% } & {\footnotesize (c)~3 Clicks~:~67.9\% } \\
\end{tabular}
\vspace{0mm}
\caption{Comparisons of scribble-based method~(ScribbleSeg) and click-based method~(FocalClick~\cite{focalclick}). }
\vspace{-3mm}
\label{fig:compare}
\end{figure}

After tuning the settings explored above, we get a strong baseline for scribble-based interactive segmentation, which already displays great performance according to Tab.~\ref{tab:evaluation sota}. In the next section, we add PAM and CRM  to make further improvements and analyze the details of these two modules.
In Tab.~\ref{tab:ablation}, we show that PAM and CRM could enhance the performance of the strong baseline independently, and combining both of them could make further improvements. 

\vspace{-4mm}
\paragraph{Prototype Adaption.} PAM gathers information from the scribble-marked regions and the mask regions to update the projection kernel. This assists ScribbleSeg in making more consistent predictions. PAM is composed of mask-pooling and scribble-pooling-guided prototype adaption. The results in Tab.~\ref{tab:PAM} demonstrate that both of these two modules are effective.

\vspace{-4mm}
\paragraph{Corrective Refine.} CRM makes detailed refinement in the predicted error regions. In Tab.~\ref{tab:CRM}, we make analyses for the different implementations. The results show that the error map is important for CRM, as it enables CRM to focus on the fine details. Detaching the feature and masks from the previous stage also brings improvements.

\begin{table}[t]
\begin{center}
\scalebox{0.7}{
\begin{tabular}{ll|c|c|c|c|c}
\toprule[1pt]
\multicolumn{2}{l|}{} & Berkeley~\cite{berkeley} & \multicolumn{2}{c|}{SBD~\cite{SBD}} & \multicolumn{2}{c}{DAVIS~\cite{davis}} \\

\multicolumn{2}{l|}{Method }  & NoI~90 & NoI~85 & NoI~90 & NoI~85 & NoI~90 \\
\hline
\multicolumn{2}{l|}{f-BRS-B-hr32~\cite{fbrs}}   & 2.44 & 4.37 & 7.26 & 5.17 & 6.50 \\
\multicolumn{2}{l|}{ RITM-hr18s~\cite{sofiiuk2021ritm}}   & 2.60 & 4.04 &  6.48 & 4.70 & 5.98 \\
\multicolumn{2}{l|}{ RITM-hr32~\cite{sofiiuk2021ritm}}     & 2.10 & 3.59 & 5.71 & 4.11 & 5.34 \\
\multicolumn{2}{l|}{ FocalClick-hr18s-S2~\cite{focalclick} }   & 2.66 & 4.43 &  6.79 & 3.90 & 5.25 \\

\multicolumn{2}{l|}{ FocalClick-B0-S2~\cite{focalclick}}   & 2.27 & 4.56 & 6.86 & 4.04 & 5.49 \\
\multicolumn{2}{l|}{ FocalClick-B3-S2~\cite{focalclick}}  &  {1.92}  & {3.53} & {5.59} & {3.61} & {4.90} \\

\hline
\multicolumn{2}{l|}{ SribbleBase-B0}   &  1.66  & 2.18 & 4.50 & 3.67 & 5.14  \\
\multicolumn{2}{l|}{ ScribbleSeg-B0}  &  1.49 & 2.56 & 4.21 & 3.29 & 4.68  \\
\rowcolor{gray!20} 
\multicolumn{2}{l|}{ ScribbleSeg-B3} & \bf1.35 & \bf2.42 & \bf3.99 & \bf3.10 & \bf4.45  \\
\bottomrule[1pt]
\end{tabular}
}
\end{center}
\vspace{-2mm}
\caption{Evaluation results on Berkeley, SBD and DAVIS datasets. 
 `NoI~85/90' denotes the average Number of Interactions~(clicks or scribbles) required the get IoU of 85/90\%.  }
\label{tab:evaluation sota}
\vspace{0mm}
\end{table}

\begin{figure*}[t]
\newcommand{\image}{\includegraphics[width=0.49\columnwidth]}
\centering 
\tabcolsep=0.04cm
\renewcommand{\arraystretch}{0.06}
\begin{tabular}{cccc}
\vspace{3pt}
\image{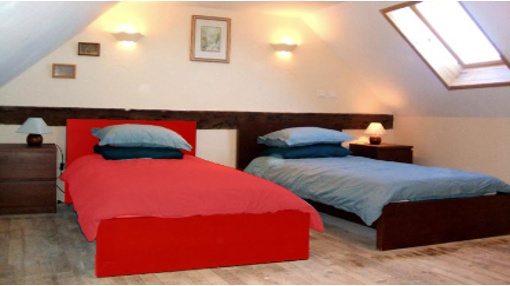} &
\image{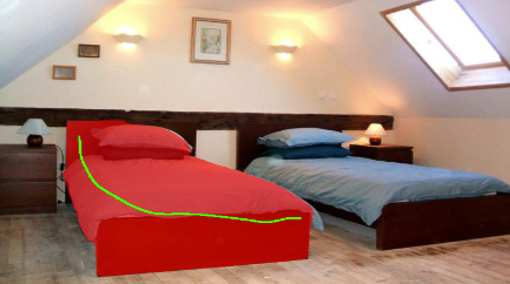} &
\image{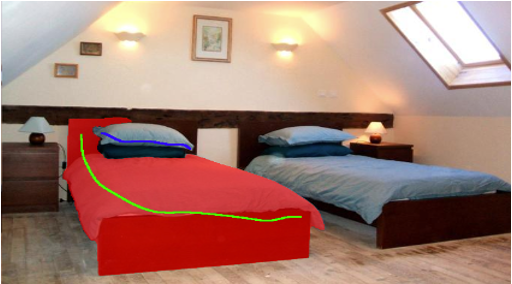} &
\image{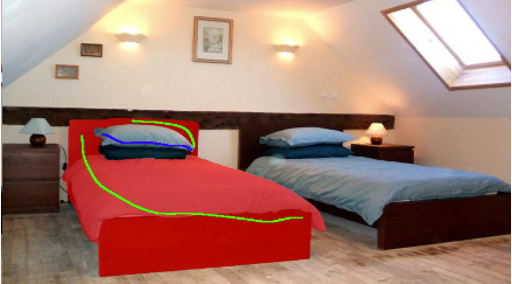} \\
\vspace{3pt}
{\footnotesize (1)~Ground Truth } &{\footnotesize (1)~1 Scribble~:~85.2\% } & {\footnotesize (1)~2 Scribbles~:~94.6\% } & {\footnotesize (1)~3 Scribbles~:~96.8\% }\\
\vspace{3pt}
\image{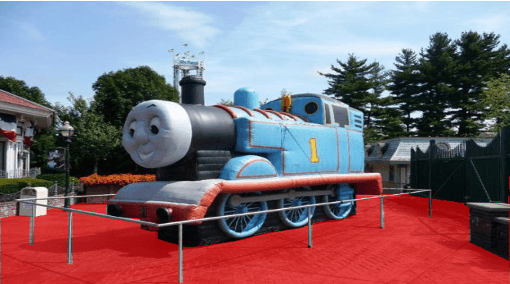} &
\image{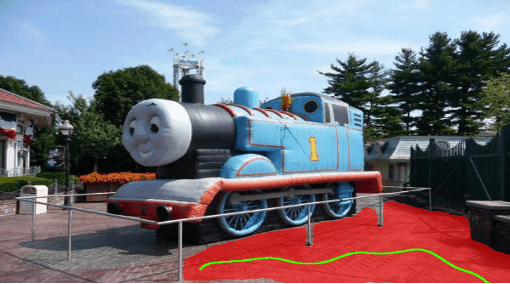} &
\image{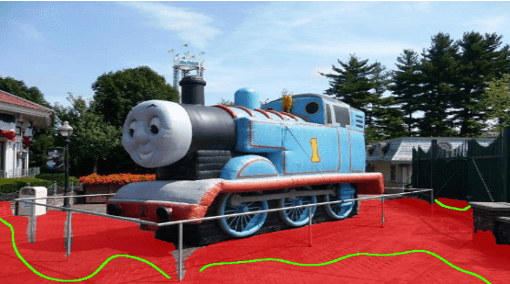} &
\image{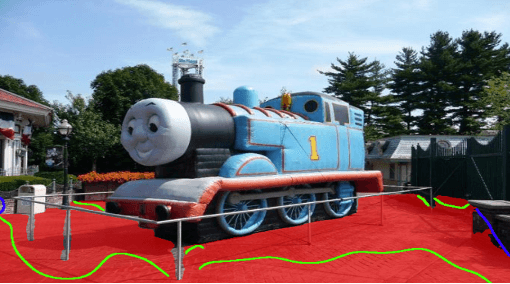} \\
\vspace{3pt}
{\footnotesize (2)~Ground Truth } &{\footnotesize (2)~1 Scribble~:~51.5\% } & {\footnotesize (2)~3 Scribbles~:~90.1\% } & {\footnotesize (2)~8 Scribbles~:~92.4\% }\\
\vspace{3pt}
\image{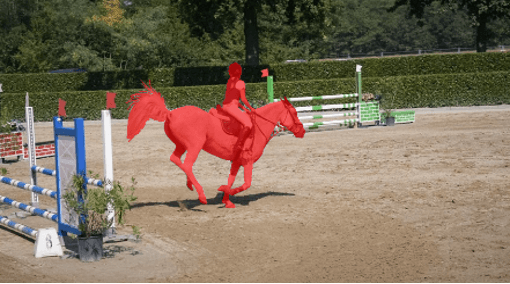} &
\image{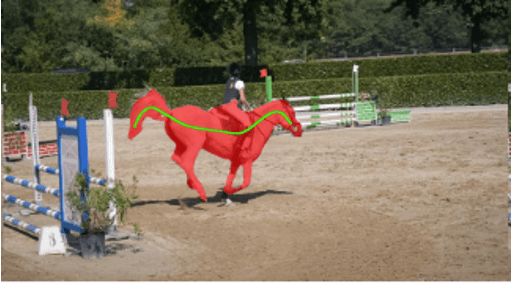} &
\image{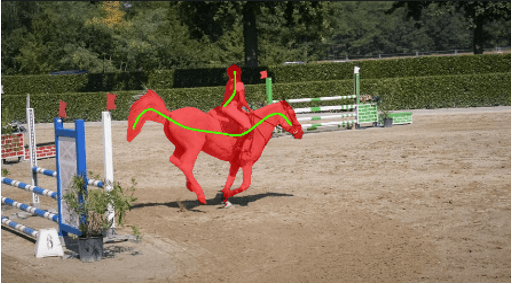} &
\image{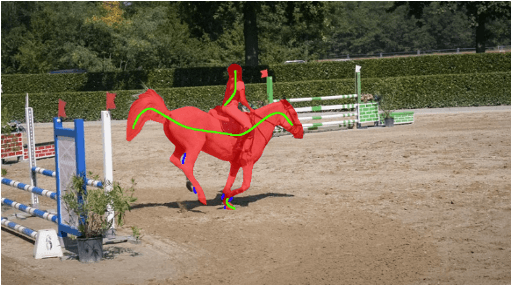} \\
\vspace{3pt}
{\footnotesize (3)~Ground Truth } &{\footnotesize (3)~1 Scribble~:~83.9\% } & {\footnotesize (3)~3 Scribbles~:~92.5\% } & {\footnotesize (3)~8 Scribbles~:~94.9\% }\\

\vspace{3pt}
\image{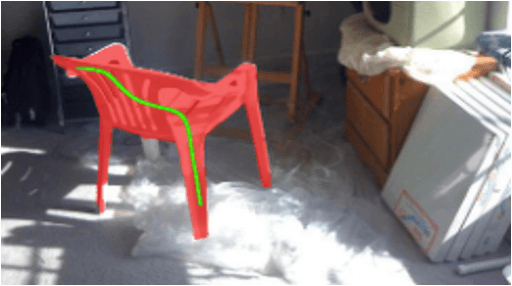} &
\image{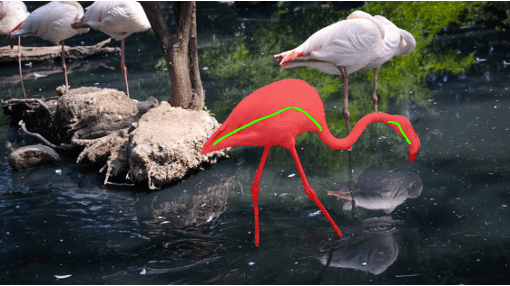} &
\image{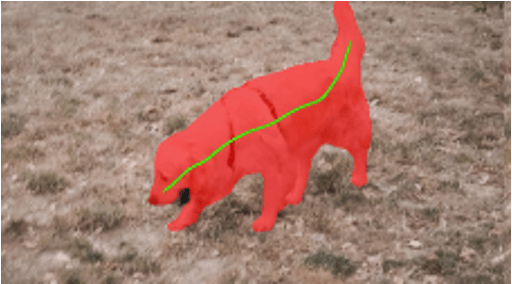} &
\image{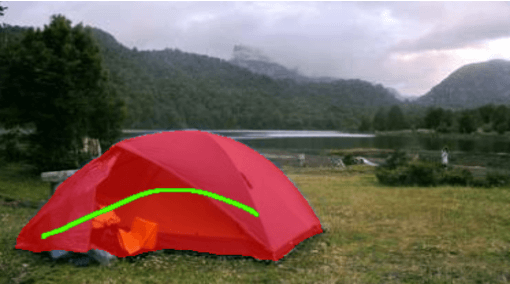} \\
\vspace{3pt}
{\footnotesize (4)~1 Scribble~:~90.2\% } &{\footnotesize (5)~2 Scribbles~:~96.7\% } & {\footnotesize (6)~1 Scribble~:~95.5\% } & {\footnotesize (7)~1 Scribble~:~95.6\% }\\

\vspace{3pt}
\image{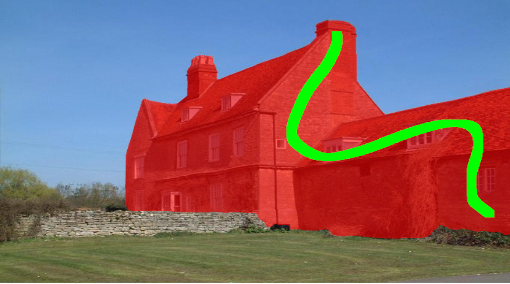} &
\image{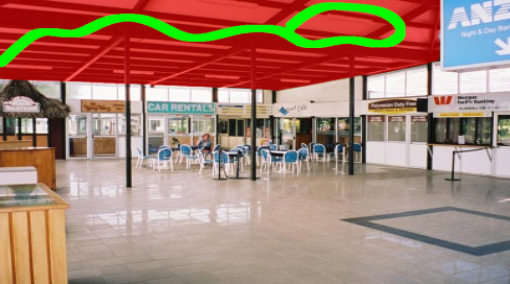} &
\image{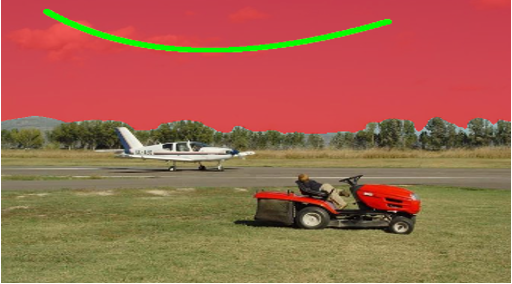} &
\image{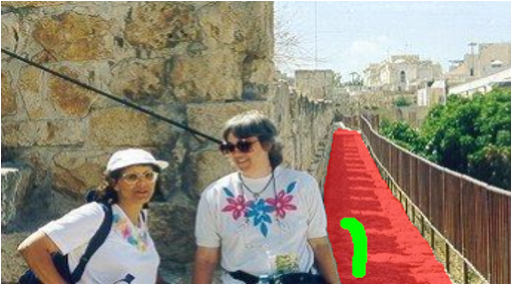} \\
\vspace{3pt}
{\footnotesize (8)~1 Scribble~:~91.4\% } &{\footnotesize (9)~2 Scribbles~:~93.4\% } & {\footnotesize (10)~1 Scribble~:~98.2\% } & {\footnotesize (11)~1 Scribble~:~94.0\% }\\

\end{tabular}
\vspace{0mm}
\caption{ Visualization results for ScribbleSeg-B3 on ADE20K~\cite{ade20k} and DAVIS~\cite{davis}. The numbers of scribbles and the IOU is marked below each image. The positive and negative scribbles are marked in green and blue. Demos 1-7 show the deterministic scribbles, which demonstrate our automatic evaluation procedure.  Demos 8-11 show the user customer scribbles. }
\vspace{-4mm}
\label{fig:demo}
\end{figure*}

\subsection{ Comparisons with Click-based Methods} 
Considering that our evaluation protocol introduced in Sec.~\ref{val_protocol} is compatible with click-based methods in a general form, we could directly compare our ScribbleSeg with previous click-based solutions. We first compare ScribbleSeg with the SOTA solutions for the click-based setting to show the benefits of using scribbles as the interaction form and prove the effectiveness of our method.

\vspace{0mm}
\paragraph{Quatitative comparisons.} In Tab.~\ref{tab:evaluation sota}, we list the click-based SOTA methods, and use our generalized metrics introduced in Sec.~\ref{val_protocol} to perform comparisons for our ScribbleSeg. We also report the performance of our strong baseline without PAM and CRM. The results show that our baseline already surpasses all click-based methods. This reflects the superiority of using scribbles as the interaction format. With PAM and CRM, the full version ScribbleSeg gets steady improvements. 

\vspace{-4mm}
\paragraph{Qualitative results.} In Fig.~\ref{fig:compare}, we compare  ScribbleSeg with the click-based SOTA method FocalClick~\cite{focalclick}. The results show that when given only one stroke of scribble, ScribbleSeg could outperform FocalClick with 3 clicks. It is clear that scribbles could provide significantly more indications than clicks. We hope ScribbleSeg could serve as a preferred choice for interactive segmentation.

\subsection{Qualitative Results}
The qualitative results for ScribbleSeg are demonstrated in Fig.~\ref{fig:demo}, where we use SegFormer-B3~\cite{xie2021segformer} as the segmentation model and make predictions on DAVIS~\cite{davis} and ADE20K~\cite{ade20k} benchmarks. 

\vspace{-4mm}
\paragraph{Evaluation procedure.}
In 1-4 rows, we show the evaluation procedure for sequentially added scribbles with the deterministic simulator. Results show that ScribbleSeg performs well on both things and stuff across diverse scenes. 

\vspace{-4mm}
\paragraph{User customer scribbles.}
The examples in the demo 8-11 show the user given scribbles with arbitrary shapes and thicknesses.
It demonstrates the robustness and generalization ability of ScribbleSeg for different interactions.
\vspace{-3pt}
\section{Limitation}
\vspace{-3pt}
Although ScribbleSeg shows great performance across different benchmarks, this version of the model would only be applicable to natural images as it is only trained on COCO~\cite{lin2014coco} and LVIS~\cite{gupta2019lvis}. If we want to use it on other domains like industrial defects and medical images, we have to collect data from the target domain and finetune the model.

\vspace{-3pt}
\section{Conclusion}
\vspace{-3pt}
We are the first to formally address the task of scribble-based interactive image segmentation. We have constructed 
the standard train/val protocol and propose ScribbleSeg. Our method shows clear advantages compared with previous click-based models. We hope this work could serve as the baseline and assist the community in making further explorations on scribble-based interactive image segmentation and developing more powerful mask annotation tools.

{\small
\bibliographystyle{ieee_fullname}
\bibliography{egbib}
}

\end{document}